\documentclass[letterpaper, 10 pt, conference]{ieeeconf}  

\IEEEoverridecommandlockouts                              

\usepackage{graphics} 
\usepackage{epsfig} 
\usepackage{times} 
\usepackage{amsmath} 
\usepackage{amssymb}  
\usepackage{booktabs}
\usepackage{bm}
\usepackage{bbm}
\usepackage{url}
\usepackage{graphicx}
\usepackage{color}
\usepackage{framed}
\usepackage{rotating}
\usepackage{multirow}
\usepackage{xspace}
\usepackage[pagebackref=true,breaklinks=true,bookmarks=false]{hyperref}
\usepackage[export]{adjustbox}

\def \ie {\emph{i.e.}}
\def \eg {\emph{e.g.}}
\def \etal {\emph{et al.}}

\def \T {{\intercal}}

\def \baseline {SMArT \textit{w/o memory}\xspace}
\def \ours {SMArT\xspace}

\title{\LARGE \bf
\textit{SMArT}: 
Training Shallow Memory-aware Transformers \\ for Robotic Explainability
}

\author{Marcella Cornia$^{1}$, Lorenzo Baraldi$^{1}$, and Rita Cucchiara$^{1}$
\thanks{$^{1}$Marcella Cornia, Lorenzo Baraldi, and Rita Cucchiara are with the Department of Engineering ``Enzo Ferrari'', University of Modena and Reggio Emilia, Modena, Italy
        {\tt\small \{name.surname\}@unimore.it}}%
}

\begin{document}

\maketitle
\thispagestyle{empty}
\pagestyle{empty}

\begin{abstract}

The ability to generate natural language explanations conditioned on the visual perception is a crucial step towards autonomous agents which can explain themselves and communicate with humans. While the research efforts in image and video captioning are giving promising results, this is often done at the expense of the computational requirements of the approaches, limiting their applicability to real contexts. In this paper, we propose a fully-attentive captioning algorithm which can provide state-of-the-art performances on language generation while restricting its computational demands. Our model is inspired by the Transformer model and employs only two Transformer layers in the encoding and decoding stages. Further, it incorporates a novel memory-aware encoding of image regions. Experiments demonstrate that our approach achieves competitive results in terms of caption quality while featuring reduced computational demands. Further, to evaluate its applicability on autonomous agents, we conduct experiments on simulated scenes taken from the perspective of domestic robots.

\end{abstract}

\section{Introduction}
Recent advancements at the intersection of Computer Vision, Natural Language Processing and Robotics have tried to bring together the understanding of the visual world and that of natural language from the perspective of robots and embodied agents~\cite{anderson2018evaluation,anderson2018vision,wang2018look,fried2018speaker,landi2019embodied}. Such research effort has the final goal of developing autonomous agents which can naturally perceive, understand, and perform actions in the surrounding world, while still having a comprehension of the human language which is fundamental to interact with the final user. The ability to describe images and videos is one of the core challenges in this domain and a crucial achievement towards machine intelligence~\cite{lu2017knowing,lu2018neural,anderson2018bottom,cornia2019show}. It requires not only to recognize salient objects in an image, understand their interactions, but also to verbalize them using natural language, which makes itself very challenging.

Noticeably, the generation of natural language conditioned on the visual input is also a critical step in the direction of cognitive system explainability and trustworthy Artificial Intelligence, as it endows an autonomous system with the ability to describe the reason of its choices, actions and to demonstrate its perception capabilities to the user. To this aim, the research efforts in language generation and image captioning are giving promising results.

Inspired by the developments in natural language processing and by the advancements in attention modeling, image captioning algorithms have relied on the combination of an image encoder and a natural language decoder. The interaction between vision and language has been modeled either using Recurrent Neural Networks or exploring more recent alternatives -- like one-dimensional convolutions or fully-attentive models such as the Transformer~\cite{vaswani2017attention,herdade2019image}. Most of the last advancements in the field, however, are due to approaches which rely on complex forms of attention and of interactions between the visual and the textual domain~\cite{yao2018exploring,jiang2018recurrent,herdade2019image}. This is often done at the expense of the computational demands of the algorithm, thus limiting the applicability of these results to embedded agents and robots. Further, for these approaches to be adapted in robotics, they need to be re-thought in terms of efficiency, memory and, power consumption as well as in terms of their adaptability in real contexts.

Following recent research lines on the investigation of fully-attentive models for image captioning~\cite{herdade2019image}, in this paper we propose a shallow and computationally efficient model for image captioning. Our model is inspired by the Transformer approach and incorporates a novel memory-aware image encoder which can model the relationships between image regions by also memorizing knowledge learned from data in a computationally friendly manner.
Further, we demonstrate state-of-the-art performances using solely two attentive layers in both the encoder and the decoder. This is in contrast with both machine translation models based on attention, and recent attempts to develop captioning systems based on the Transformer, which tend to use six or more encoding and decoding layers. Our approach is competitive in terms of caption quality and has the additional benefit of having reduced computational demands when compared with recent approaches.

Summarizing, our contributions are as follows: (i) we introduce \textit{SMArT}, a Shallow and Memory-Awa\textit{r}e Transformer for the task of image captioning; (ii) our model incorporates self-attention and a memory-aware image encoding layer, in which self-attention is endowed with memory vectors; (iii) we demonstrate competitive results on the reference benchmark for image captioning (COCO) using only two encoder and two decoder layers. Finally, we demonstrate the applicability of our approach to simulated scenes taken from the perspective of domestic robots.

\section{Related Work} \label{sec:related}
A large variety of methods has been proposed for the image captioning task. While early approaches were based on caption templates filled by using the output of pre-trained object detectors~\cite{yao2010i2t,socher2010connecting}, almost all recent captioning models integrate recurrent neural networks as language models~\cite{lu2017knowing,liu2017improved,johnson2016densecap,lu2018neural} with a visual feature extractor for conditioning the language model on the visual input.

The representation of the image has been initially obtained from the output of one or more layer of a CNN~\cite{vinyals2017show,donahue2015long,rennie2017self,lu2017knowing}. Then, with the integration of attentive mechanisms~\cite{bahdanau2014neural}, the visual representation has turned into a time-varying vector extracted from a grid of CNN features using the hidden state of the language model as the query~\cite{xu2015show,you2016image,lu2017knowing,cornia2018paying,anderson2018bottom}. Recently, integrating image regions eventually extracted from a detector as attention candidates has become the predominant strategy in captioning architectures~\cite{pedersoli2016areas,anderson2018bottom}. On this line, Jiang~\etal~\cite{jiang2018recurrent} proposed a recurrent fusion network to integrate the output of multiple image encoders. Yao~\etal~\cite{yao2018exploring} have explored the incorporation of relationships between image regions, both from a semantic point of view and using geometric features such as the position and the spatial extent of the region. Their work exploits semantic relationships predictor which are trained separately, and whose outputs are incorporated inside a Graph Convolutional Neural Network. 
Regarding the training strategies, notable advances have been made by using Reinforcement Learning to train non-differentiable captioning metrics~\cite{ranzato2015sequence,liu2017improved,rennie2017self}. 

\begin{figure*}[t]
\centering
  \includegraphics[width=\textwidth]{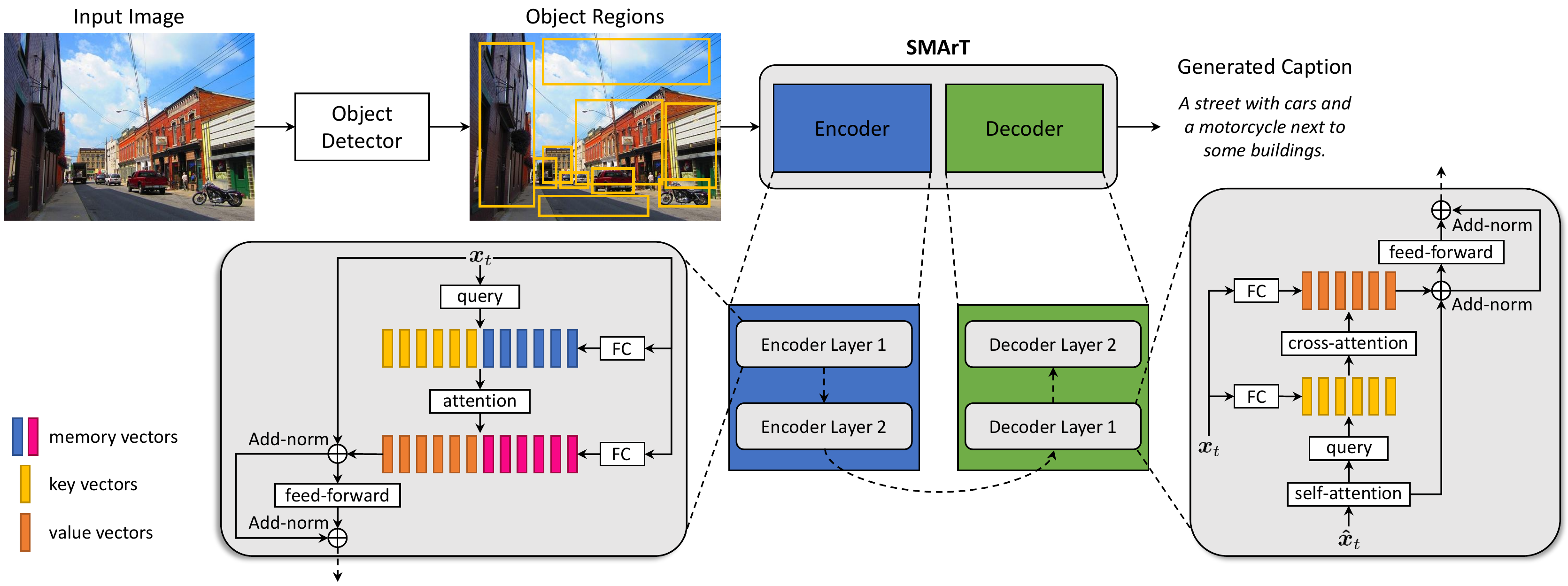}
  \caption{Overview of our image captioning approach. Building on a Transformer-like encoder-decoder architecture, our approach includes a memory-aware region encoder which augments self-attention with memory vectors. Further, our approach is shallow, as it requires only two encoding and decoding layers.}
  \label{fig:method}
  \vspace{-0.1cm}
\end{figure*}

While RNNs have been a popular choice for defining the language model, they also suffer from their sequential nature and limited representation power. For this reason, researchers have also investigated the use of alternatives which have demonstrated to be compelling for machine translation. In this context, Aneja~\etal~\cite{aneja2018convolutional} have defined a fully-convolutional captioning model using one-dimensional convolution and attention over image features. More recently, the Transformer model~\cite{vaswani2017attention} has been proposed as a powerful and fast language model for translation tasks and unsupervised pre-training~\cite{devlin2018bert}. Herdade~\etal~\cite{herdade2019image} have first investigated the usage of the Transformer model for image captioning, proposing a deep architecture which is conditioned on geometric region features. In contrast to this concurrent work, we propose a shallow architecture based on the Transformer and the use of a novel memory-aware region encoder. When compared to this proposal, our approach demonstrates to be both less computationally demanding and more accurate in terms of CIDEr.

\section{Proposed Method}
Our fully-attentive captioning model consists of an encoder module, in charge of encoding image regions, and a decoder module, which is conditioned on the encoder and generates the natural language description. In contrast to previous captioning approaches, which employed RNNs as the language model, we  propose to use a fully-attentive model for both the encoding and the decoding stage, building on the Transformer model~\cite{vaswani2017attention} for machine translation. In addition, we propose a self-attention region encoder with memory vectors to encode learned knowledge and relationships on the visual input. A summary of our approach and of its components is visually presented in Fig.~\ref{fig:method}.

\subsection{Transformer layers}
Both encoder and decoder consist of a stack of Transformer layers which act, respectively, on image regions and words. In the following, we revise their fundamental features. Each encoder layer consists of a self-attention and feed-forward layer, while each decoder layer is a stack of one self-attentive and one cross-attentive layer, plus a feed-forward layer. Both attention layers and feed-forward layers are encapsulated into ``add-norm'' operations, described in the following.

\paragraph{Multi-head attention} the core component of both self-attention and cross-attention layers is an attention mechanism~\cite{bahdanau2014neural} with multiple heads with different learned weights. Attention is applied using scaled dot-products as similarity measure~\cite{vaswani2017attention} while keys, queries, and values are computed through linear transformations. 

Formally, given two sequences $\bm{x}_1, \bm{x}_2, ..., \bm{x}_N$ and $\bm{\hat{x}}_1, \bm{\hat{x}}_2, ..., \bm{\hat{x}}_M$ of $d$-dimensional input vectors, each head applies two linear transformations to the first sequence to form key and value vectors:
\begin{equation}
    \bm{k}_t = \bm{W}_k \bm{x}_t, \quad \bm{v}_t = \bm{W}_v \bm{x}_t,
    \label{eq:keyvalues}
\end{equation}
where $\bm{W}_k$ and $\bm{W}_v$ are the key and value transformation matrices, with size $d_h \times d$, where $d_h = d / H$ is the dimensionality of a single head, and $H$ is the number of heads.
Analogously, a linear transformation is applied to the second sequence to obtain query vectors:
\begin{equation}
    \bm{q}_t = \bm{W}_q \bm{\hat{x}}_t,
\end{equation}
where $\bm{W}_q$ has the same size of $\bm{W}_k$ and $\bm{W}_v$. Query vectors are used to compute a similarity score with key vectors, and generate a weight distribution over values. The similarity score between a query $\bm{q}_t$ and a key $\bm{k}_c$ is computed as a scaled dot-product between the two vectors, \ie~$(\bm{q}_t^\T\bm{v}_c) / \sqrt{d_h}$.
Each head then produces a vector by averaging the values $\{\bm{v}_c\}_c$ with the weights defined by an attentive distribution over the similarity scores:
\begin{align}
    \bm{y}_t &= \sum_{c=1}^N \alpha_{tc} \bm{v}_c, \text{ where} \\
    \alpha_{tc} &= \frac{\exp{(\bm{q}_t^\T\bm{v}_c / \sqrt{d_h})}}{\sum_i{\exp{(\bm{q}_t^\T\bm{v}_i / \sqrt{d_h})}}}.
\end{align}
Results from different heads are then concatenated and projected to a vector with dimensionality $d$ through a final linear transformation.

In the encoding stage, the sequence of image regions is used to infer queries, keys and values, thus creating a self-attention pattern in which pairwise region relationships are modelled. In the decoder, we instead apply both a cross-attention and a masked self-attention pattern. In the former, the sequence of words is used to infer queries and image regions are used as keys and values. In the latter, the left-hand side of the textual sequence is used to generate keys and values for each element of the sequence, thus enforcing causality in the generation.

\paragraph{Position-wise feed-forward layers} the second component of a Transformer layer is a fully-connected forward layer which is applied time-wise over the input sequence. This consists of two affine transformations with a single non-linearity,
\begin{equation}
    \texttt{FF}(\mathbf{x}_t) = \mathbf{U}\sigma(\mathbf{V} \mathbf{x}_t + \mathbf{b}) + \mathbf{c},
\end{equation}
where $\sigma(x)=\max(x, 0)$ is the \textsc{ReLU} activation function, and $\bm{V}$ and $\bm{U}$ are learnable weight matrices, respectively with sizes $d \times d_f$ and $d_f \times d$; $\mathbf{b}$ and $\mathbf{c}$ are bias terms. The size of the hidden layer $d_f$ is usually chosen to be larger than $d$, \eg~four times $d$ in most implementations~\cite{vaswani2017attention}.

\paragraph{Skip connection and layer normalization}
Each sublayer (attention or position-wise feed-forward) is encapsulated within a residual connection~\cite{he2016deep} and layer normalization~\cite{ba2016layer}. This ``add-norm'' operation is defined as
\begin{equation}
    \texttt{AddNorm}(\mathbf{x}_t) = \texttt{LayerNorm}(\bm{x}_t + f(\bm{x}_t)),
\end{equation}
where $f$ indicates either an attention layer or a position-wise feed-forward layer.

\subsection{Memory-augmented region encoder}
Recent captioning literature has demonstrated that regions identified by an object detector are the ideal attention candidates for encoding the visual input~\cite{anderson2018bottom}. A stack of self-attentive layers can naturally model the relationships between regions; however, it can not naturally encode a-priori knowledge learned from data. To overcome this limitation and enhance the visual encoding capabilities of the model, we endow our encoder with memory slots. In practice, we extend each self-attention layer of the encoder so that the key and value sets of each head contain an additional set of learned parameters.

Formally, the set of keys $\bm{K}$ of each head is extended as follows:
\begin{equation}
    \bm{K} = \left[ \underbrace{\bm{k}_1, \bm{k}_2, ..., \bm{k}_N}_{\text{ordinary keys}}, \underbrace{\bm{k}^m_1, \bm{k}^m_2, ..., \bm{k}^m_M}_{\text{memory key vectors}}\right],
\end{equation}
where the ordinary keys $\{ \bm{k}_i \}_i$ are computed through linear transformations from the sequence (Eq.~\ref{eq:keyvalues}), and memory key vectors $\{ \bm{k}^m_i \}_i$ are learnable weights that act as memory slots. As it can be seen, memory slots are independent on the input sequence of detections, and therefore store knowledge which does not depend on the input or the context.

Similarly, the values $\bm{V}$ of an head are extended with learnable memory slots as well,
\begin{equation}
    \bm{V} = \left[ \underbrace{\bm{v}_1, \bm{v}_2, ..., \bm{v}_N}_{\text{ordinary values}}, \underbrace{\bm{v}^m_1, \bm{v}^m_2, ..., \bm{v}^m_M}_{\text{memory value vectors}}\right],
\end{equation}
where $\{ \bm{v}^m_i \}_i$ are learnable weights with the same dimensionality of a value. By defining memory keys and memory values as separate weights, we break the linear dependency between keys and values, thus letting the network learn unrelated sets of keys and values.  This is in contrast with concurrent approaches in machine translation which have investigated the use of persistent memories~\cite{sukhbaatar2019augmenting}. Given $M$ as the number of key and value memory slots for each head, our model overall learns a set of $2M\cdot H$ memory slots.

\subsection{Fully-attentive decoder}
The language model of our approach is composed of a stack of two decoder layers, each performing self-attention and cross-attention operations. As mentioned, each cross-attention layer uses the decoder output sequence to infer keys and values, while self-attention layers rely exclusively on the input sequence of the decoder. However, keys and values are masked so that each query can only attend to keys obtained from previous words, \ie~the set of keys and values for query $\bm{q}_t$ are, respectively, $\{ \bm{k}_i \}_{i\le t}$ and $\{ \bm{v}_i \}_{i\le t}$.

At training time, the input of the encoder is the ground-truth sentence $\{\text{BOS}, w_1, w_2, ..., w_n\}$, and the model is trained with a cross-entropy loss to predict the shifted ground-truth sequence, \ie~$\{ w_1, w_2, ..., w_n, \text{EOS} \}$, where $\text{BOS}$ and $\text{EOS}$ are special tokens to indicate the start and the end of the caption.

While at training time the model jointly predicts all output tokens, the generation process at prediction time is sequential. At each iteration, the model is given as input the partially decoded sequence; it then samples the next input token from its output probability distribution, until a $\text{EOS}$ marker is generated.

Following previous works~\cite{ranzato2015sequence,rennie2017self,anderson2018bottom}, after a pre-training step using cross-entropy, we further optimize the sequence generation using Reinforcement Learning. Specifically, we employ a variant of the self-critical sequence training approach~\cite{rennie2017self} which applies the REINFORCE algorithm on sequences sampled using Beam Search~\cite{anderson2018bottom}. Further, we baseline the reward using the mean of the rewards rather than greedy decoding as done in~\cite{rennie2017self,anderson2018bottom}.

Specifically, given the output of the decoder we sample the top-$k$ words from the decoder probability distribution at each timestep, and always maintain the top-$k$ sequences with highest probability. We then compute the reward of each sentence $\bm{w}^i$ and backpropagate with respect to it. The final gradient expression for one sample is thus:
\begin{equation}
    \nabla_\theta L(\theta) = -\frac{1}{k}\sum_{i=1}^k \left((r(\bm{w}^i)-b) \nabla_\theta \log p(\bm{w}^i)\right)
\end{equation}
where $b = \left(\sum_i r(\bm{w}^i)\right)/k$ is the baseline, computed as the mean of the rewards obtained by the sampled sequences. 

To reward the overall quality of the generated caption, we use image captioning metrics as a reward. Following previous works~\cite{anderson2018bottom}, we employ the CIDEr metric (specifically, the CIDEr-D score) which has been shown to correlate better with human judgment~\cite{vedantam2015cider}.

\section{Experimental Evaluation}
\begin{table}[t]
\small
\centering
\caption{Captioning performance (without memory) as we vary the number of encoder and decoder layers.}
\label{tab:layers}
\setlength{\tabcolsep}{.45em}
\begin{tabular}{cccccccc}
\toprule
N. Layers & & B-1 & B-4 & M & R & C & S \\
\midrule
1       & & 80.1 & 38.1 & 28.7 & 58.0 & 127.9 & 22.7 \\
2       & & 80.2 & 38.0 & 28.9 & 58.3 & \textbf{128.9} & 22.4  \\
3       & & 79.4 & 36.4 & 27.7 & 56.8 & 123.7 & 20.9 \\
4       & & 79.2 & 36.3 & 28.0 & 57.1 & 121.7 & 21.6 \\
5       & & 79.2 & 36.2 & 27.5 & 56.7 & 120.5 & 20.6 \\
6 (as~\cite{vaswani2017attention,herdade2019image})       & & 79.1 & 36.1 & 27.9 & 56.8 & 121.9 & 21.0  \\
\bottomrule
\end{tabular}
\end{table}

\subsection{Datasets}
For comparison with the state of the art, we employ the Microsoft COCO dataset, which contains $123\,287$ images labeled with $5$ captions each. We employ the data splits defined in~\cite{karpathy2015deep}, where $5\,000$ images are used for validation, $5\,000$ images for testing and the rest for training. 
Further, to assess the performance of our approach on images taken from a robot-centric point of view, we employ the ACVR Robotic Vision Challenge dataset~\cite{hall2018probability} which contains simulated data from a domestic robot scenario. The dataset contains scenes with cluttered surfaces, and day and night lighting conditions. Authors have simulated domestic service robots of multiple sizes, resulting in sequences with three different camera heights above the ground plane. We employ the validation set of this dataset, for which ground-truth object information is available. This consists of over $21\,000$ images in four simulated indoor video sequences, containing a subset of COCO classes.

\begin{table}[t]
\small
\centering
\caption{The performance of our model as we vary the number of memory slots for each head.}
\label{tab:memories}
\setlength{\tabcolsep}{.45em}
\begin{tabular}{cccccccc}
\toprule
N. Memories & & B-1 & B-4 & M & R & C & S \\
\midrule
No memory   & & 80.2 & 38.0 & 28.9 & 58.3 & 128.9 & 22.4 \\
20       & & 80.7 & 38.5 & 28.9 & 58.2 & 129.4 & 22.4  \\
40       & & 80.4 & 38.1 & 28.8 & 58.2 & \textbf{129.7} & 22.2 \\
100       & & 80.7 & 38.5 & 29.0 & 58.4 & 128.6 & 22.6 \\
\bottomrule
\end{tabular}
\end{table}

\subsection{Evaluation protocols}
Regarding evaluation, we employ popular captioning metrics whenever ground-truth captions are available, to evaluate both fluency and semantic correctness: BLEU~\cite{papineni2002bleu}, $\text{ROUGE}$~\cite{lin2004rouge}, METEOR~\cite{banerjee2005meteor}, and CIDEr~\cite{vedantam2015cider}. BLEU is a form of precision of word n-grams
between predicted and ground-truth sentences. As done in previous works, we evaluate our predictions with BLEU using n-grams of lenght 1 and 4. $\text{ROUGE}$ computes an F-measure with a recall bias using a longest common subsequence technique. METEOR, instead, scores captions by aligning them to one or more ground-truths. Alignments are based on exact, stem, synonym, and paraphrase matches between words and phrases. CIDEr, finally, computes the average cosine similarity between n-grams found in the generated caption and those found in reference sentences, weighting them using TF-IDF.  While it has been shown experimentally that BLEU and ROUGE have lower correlation with human judgments than the other metrics~\cite{vedantam2015cider}, the common practice in the image captioning literature is to report all the mentioned metrics. To ensure a fair evaluation, we use the Microsoft COCO evaluation toolkit to compute all scores.

When only object-level information is available as ground-truth, such as in the ACVR dataset, we evaluate the capability of our captioning approach to name objects on the scene.
To assess how the predicted caption covers all the objects, we also define a soft coverage measure between the ground-truth set of object classes and the set of names in the caption. Given a predicted caption $\bm{y}$, we firstly extract all nouns from the sentence. We compute the optimal assignment between them and the set of ground-truth classes $\bm{c}^*$, using distances between word vectors and the Hungarian algorithm~\cite{kuhn1955hungarian}. We then define an intersection score between the two sets as the sum of assignment profits. Our coverage measure is computed as the ratio of the intersection score and the number of ground-truth object classes:
\begin{equation}
    \text{Cov}(\bm{y}, \bm{c}^*) = \frac{\text{I}(\bm{y}, \bm{y}^*)}{\#\bm{c}^*},
\end{equation}
where $\text{I}(\cdot, \cdot)$ is the intersection score, and the $\#$ operator represents the cardinality of the two sets of nouns.

Since images may contain small objects which not necessarily should be mentioned in a caption describing the overall scene, we also define a variant of the Coverage measure by thresholding over the minimum object area. In this case, we consider $\bm{c}^*$ as the set of objects whose bounding boxes cover an area higher than the threshold.

\subsection{Implementation and training details}
As mentioned, we use two layers in both the encoder and the decoder. The dimensionality of all layers, $d$, is set to 512 and we use $H=8$ heads. The dimensionality of the inner feed-forward layer, $d_f$, is 2048. We use dropout with keep probability 0.9 after each attention layer and after position-wise feed-forward layers. Input words are represented with one-hot vectors and then linearly projected to the input dimensionality of the model, $d$. We also employ sinusoidal positional encodings~\cite{vaswani2017attention} to represent word positions inside the sequence, and sum the two embeddings before the first encoding layer.

To represent image regions, we use Faster R-CNN~\cite{ren2015faster} with ResNet-101~\cite{he2016deep}. In particular, we employ the model finetuned on the Visual Genome dataset~\cite{krishnavisualgenome} provided by~\cite{anderson2018bottom}. To compute the intersection score and for extracting nouns from captions, we use the spaCy NLP toolkit\footnote{\url{https://spacy.io/}}. We use GloVe~\cite{pennington2014glove} as word vectors.

The model is trained using Adam~\cite{kingma2014adam} as optimizer with $\beta_1=0.9$ and $\beta_2=0.98$. The learning rate is varied during training using the strategy of~\cite{vaswani2017attention}, \ie~according to the formula: $d^{-0.5}\cdot \min(s^{-0.5}, s\cdot w^{-0.5})$, where $s$ is the current optimization step and $w$ is a warmup parameter, set to $10\,000$ in all our experiments. After the pre-training with cross-entropy loss, we finetune the model again using Adam and with a fixed learning rate of $5e^{-6}$.

\begin{table}[t]
\small
\centering
\caption{Comparison with the state of the art for image captioning on the test set of the COCO dataset. }
\label{tab:sota_results}
\setlength{\tabcolsep}{.45em}
\begin{tabular}{lccccccc}
\toprule
Method & & B-1 & B-4 & M & R & C & S  \\
\midrule
FC-2K~\cite{rennie2017self} & & - & 31.9 &  25.5 & 54.3 & 106.3 & - \\
Att2all~\cite{rennie2017self} & & - & 34.2 & 26.7 & 55.7 & 114.0 & - \\
Up-Down~\cite{anderson2018bottom}  & & 79.8 & 36.3 & 27.7 & 56.9 & 120.1 & 21.4 \\
RFNet~\cite{jiang2018recurrent} & & 79.1 & 36.5 & 27.7 & 57.3 & 121.9 & 21.2 \\
SGAE~\cite{yang2019auto} & & 80.8 & 38.4 & 28.4 & 58.6 & 127.8 & 22.1 \\ 
GCN-LSTM~\cite{yao2018exploring}  & & 80.9 & 38.3 & 28.6 & 58.5 & 128.7 & 22.1 \\
ORT~\cite{herdade2019image} & & 80.5 & 38.6 & 28.7 & 58.4 & 128.3 & 22.6 \\
\midrule
\textbf{\baseline}        & & 80.2 & 38.0 & 28.9 & 58.3 & 128.9 & 22.4 \\
\textbf{\ours} ($m=40$)   & & 80.4 & 38.1 & 28.8 & 58.2 & \textbf{129.7} & 22.2 \\
\bottomrule
\end{tabular}
\end{table}

\begin{table}[t]
\small
\centering
\caption{Object coverage analysis on the ACVR Robotic Vision Challenge dataset, when varying the minumum object area threshold.}
\label{tab:acrv}
\setlength{\tabcolsep}{.5em}
\begin{tabular}{lccccc}
\toprule
& & \multicolumn{4}{c}{Coverage}    \\
\cmidrule{3-6}
Method & & $>1\%$ & $>3\%$ & $>5\%$ & $>10\%$  \\
\midrule
\baseline       & & 0.747 & 0.806 & 0.836 & 0.846 \\
\ours ($m=20$)  & & 0.751 & 0.808 & 0.841 & 0.846 \\
\ours ($m=40$)  & & \textbf{0.762} & \textbf{0.821} & \textbf{0.848} & \textbf{0.850} \\
\ours ($m=100$) & & 0.757 & 0.814 & 0.843 & 0.846 \\
\bottomrule
\end{tabular}
\end{table}

\subsection{Captioning results}
\subsubsection{Shallow vs. deep models}
Firstly, we investigate the performance of fully-attentive captioning models as we vary the number of encoding and decoding layers. In particular, we start from our model without memory, and keep the number of encoding and decoding layers equal. Table~\ref{tab:layers} shows the results obtained after a full training with RL finetuning. Noticeably, the performance obtained with six layers (as in the original Transformer model~\cite{vaswani2017attention}, and as in the captioning model of~\cite{herdade2019image}) is lower than the one obtained when using 1, 2 or 3 layers. While the best results are obtained with two layers (128.9 CIDEr), we notice that using just one layer is a compelling alternative, which still obtains a CIDEr of 127.9. 

\subsubsection{Persistent memory vectors}
We then evaluate the role of using persistent memory vectors in the encoder. Table~\ref{tab:memories} reports the performance obtained by our model with two layers and a number of memory slots per head varying from 0 to 100. As it can be seen, using 40 memory slots for each head further increases the CIDEr metric from 128.9 to 129.7. 

\begin{figure*}[t]
\centering
\includegraphics[width=\linewidth]{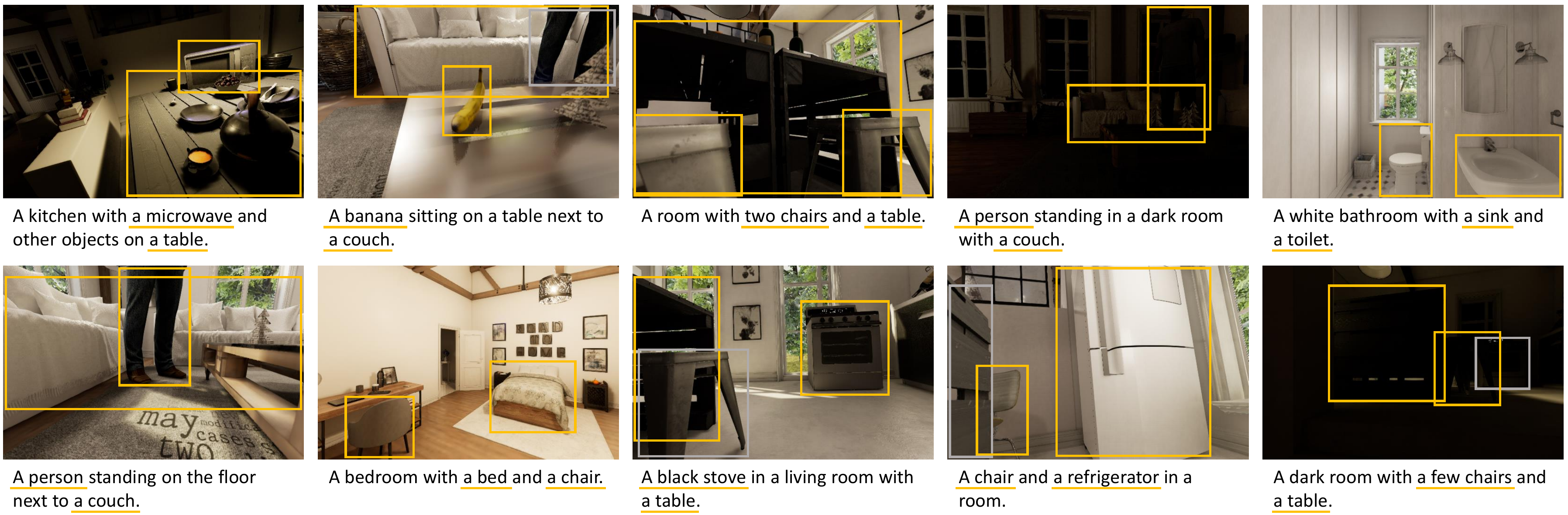}
\caption{Sentences generated on the ACVR Robotic Vision Challenge dataset. We report objects detected on the scene and underline their mentions in the caption. Objects present in the scene and not mentioned in the caption are shown in gray.}
\label{fig:acrv}
\end{figure*}

\begin{figure}[t]
\centering
  \includegraphics[width=.85\columnwidth]{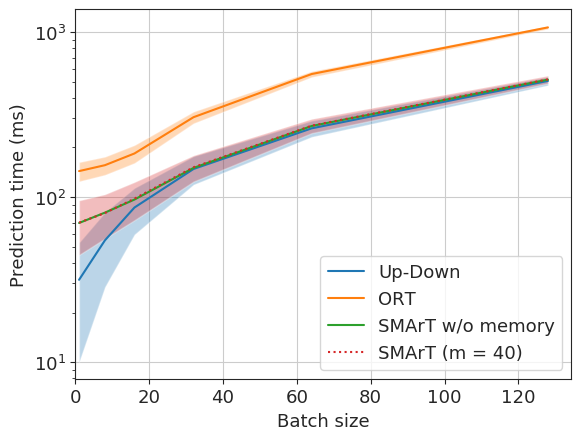}
  \caption{Prediction times when varying the batch size, for SMArT (w/ and w/o memory), ORT, and Up-Down. SMArT features higher fluency and correctness with comparable prediction times compared to LSTM-based approaches; lower execution times and comparable correctness when compared to Transformer-based approaches.}
  \label{fig:performance}
\end{figure}

\subsubsection{Comparison with the state of the art}
We report the performances of our model in comparison with different captioning models. In particular, we compare with: FC-2K~\cite{rennie2017self}, an LSTM baseline using a global feature vector as image descriptor; Att2all~\cite{rennie2017self}, which uses additive attention over the grid of image features extracted from a CNN; 
Up-Down~\cite{anderson2018bottom}, which employs a two-LSTM layer model with bottom-up features extracted from Faster R-CNN. Also, we compare with RFNet~\cite{jiang2018recurrent}, which fuses encoded features from multiple CNN networks; SGAE~\cite{yang2019auto}, which introduces auto-encoding scene graphs into its model. Finally, we also consider GCN-LSTM~\cite{yao2018exploring}, which explores the role of visual relationships between image regions and, importantly, with ORT~\cite{herdade2019image}, which employed a Transformer-based model with six layers.

Table~\ref{tab:sota_results} reports the performances of all the mentioned approaches at the end of training with reinforcement learning. Firstly, we notice that our approach overcomes both LSTM-based approaches and ORT~\cite{herdade2019image} according to the CIDEr metric. Noticeably also, the performance of a shallow model without memory is in line with the state of the art. On all the other metrics, which are less aligned with human judgment, our approach achieves competitive performances.

\subsubsection{Performance on robot-centric images}
Table~\ref{tab:acrv} reports the Coverage measure for different variants of our approach on the validation set of the ACVR Robotic Vision Challenge dataset. As it can be observed, our approach is capable of mentioning most of the objects on the scene which cover at least $10\%$ of the image, and achieves a 0.76 coverage score when thresholding at $1\%$ of the area, thus taking into account also small objects. Reported results also confirm the findings on the effectiveness of the persistent memory vectors observed on COCO. To further highlight the adaptability of our solution to robot-centric images, Figure~\ref{fig:acrv} shows some predicted captions on sample images of the ACVR dataset.

\subsection{Computational analysis}
We complete our experimental evaluation by analyzing the computational demands of our solution in comparison with ORT~\cite{herdade2019image} and Up-Down~\cite{anderson2018bottom}. We compare with ORT~\cite{herdade2019image} as it is the only Transformer-based competitor, and with Up-Down~\cite{anderson2018bottom} as it is the most lightweight among recent proposals based on LSTM. 
Adapt-Att~\cite{lu2017knowing}, RFNet~\cite{jiang2018recurrent} and GCN-LSTM~\cite{yao2018exploring} can indeed be seen as extensions of Up-Down~\cite{anderson2018bottom} which add elements to its computational graph, thus potentially increasing processing times.

Figure~\ref{fig:performance} shows the mean and standard deviation of prediction times as a function of the batch size. For a fair evaluation, we run all approaches on the same workstation and on the same GPU (NVIDIA 1080Ti). To exclude the effect of different implementations, we re-implement all the approaches and use the same framework (\ie~PyTorch) and the same routines whenever possible. As it can be observed, SMArT is more computationally efficient than ORT~\cite{herdade2019image}, reducing the prediction times by a significant margin. Also, the use of persistent memories does not significantly impact prediction performance. When comparing with Up-Down, instead, we notice that the mean prediction time is lower for small batch sizes, while for larger batch sizes the performances of Up-Down and of our approach are comparable. Considering the captioning quality given by Up-Down (120.1 CIDEr) and that of our approach (129.7 CIDEr), our method provides better caption quality with a comparable computational cost. Most importantly, the reduction in the number of layers provides better caption quality (129.7 CIDEr vs. 128.3 CIDEr) and reduced prediction times compared to other Transformer-based approaches.

\section{Conclusion}
In this paper we have presented SMArT, a novel approach for image captioning, based on a shallow Transformer network endowed with persistent memory vectors. Our solution is motivated by the need of effective bridges between vision, language and action that can be deployed on autonomous agents. As shown in the experimental evaluation, SMArT achieves state of the art caption quality with reduced computational needs. Additionally, we have demonstrated the applicability of our approach to robot-centric images from simulated environments.





\section*{Acknowledgment}
This work was partially supported by the ``IDEHA - Innovation for Data Elaboration in Heritage Areas'' project (PON ARS01\_00421), funded by the Italian Ministry of Education (MIUR). We also gratefully acknowledge the NVIDIA AI Technology Center, EMEA, for its support and access to computing resources.


\bibliographystyle{IEEEtran}
\bibliography{bibliography}

\begin{thebibliography}{10}
\providecommand{\url}[1]{#1}
\csname url@rmstyle\endcsname
\providecommand{\newblock}{\relax}
\providecommand{\bibinfo}[2]{#2}
\providecommand\BIBentrySTDinterwordspacing{\spaceskip=0pt\relax}
\providecommand\BIBentryALTinterwordstretchfactor{4}
\providecommand\BIBentryALTinterwordspacing{\spaceskip=\fontdimen2\font plus
\BIBentryALTinterwordstretchfactor\fontdimen3\font minus
  \fontdimen4\font\relax}
\providecommand\BIBforeignlanguage[2]{{%
\expandafter\ifx\csname l@#1\endcsname\relax
\typeout{** WARNING: IEEEtran.bst: No hyphenation pattern has been}%
\typeout{** loaded for the language `#1'. Using the pattern for}%
\typeout{** the default language instead.}%
\else
\language=\csname l@#1\endcsname
\fi
#2}}

\bibitem{anderson2018evaluation}
P.~Anderson, A.~Chang, D.~S. Chaplot, A.~Dosovitskiy, S.~Gupta, V.~Koltun,
  J.~Kosecka, J.~Malik, R.~Mottaghi, M.~Savva, \emph{et~al.}, ``On evaluation
  of embodied navigation agents,'' \emph{arXiv preprint arXiv:1807.06757},
  2018.

\bibitem{anderson2018vision}
P.~Anderson, Q.~Wu, D.~Teney, J.~Bruce, M.~Johnson, N.~S{\"u}nderhauf, I.~Reid,
  S.~Gould, and A.~van~den Hengel, ``Vision-and-language navigation:
  Interpreting visually-grounded navigation instructions in real
  environments,'' in \emph{Proceedings of the IEEE Conference on Computer
  Vision and Pattern Recognition}, 2018.

\bibitem{wang2018look}
X.~Wang, W.~Xiong, H.~Wang, and W.~Yang~Wang, ``Look before you leap: Bridging
  model-free and model-based reinforcement learning for planned-ahead
  vision-and-language navigation,'' in \emph{Proceedings of the European
  Conference on Computer Vision}, 2018.

\bibitem{fried2018speaker}
D.~Fried, R.~Hu, V.~Cirik, A.~Rohrbach, J.~Andreas, L.-P. Morency,
  T.~Berg-Kirkpatrick, K.~Saenko, D.~Klein, and T.~Darrell, ``Speaker-follower
  models for vision-and-language navigation,'' in \emph{Advances in Neural
  Information Processing Systems}, 2018.

\bibitem{landi2019embodied}
F.~Landi, L.~Baraldi, M.~Corsini, and R.~Cucchiara, ``{Embodied
  Vision-and-Language Navigation with Dynamic Convolutional Filters},'' in
  \emph{Proceedings of the British Machine Vision Conference}, 2019.

\bibitem{lu2017knowing}
J.~Lu, C.~Xiong, D.~Parikh, and R.~Socher, ``Knowing when to look: Adaptive
  attention via a visual sentinel for image captioning,'' in \emph{Proceedings
  of the IEEE Conference on Computer Vision and Pattern Recognition}, 2017.

\bibitem{lu2018neural}
J.~Lu, J.~Yang, D.~Batra, and D.~Parikh, ``{Neural Baby Talk},'' in
  \emph{Proceedings of the IEEE Conference on Computer Vision and Pattern
  Recognition}, 2018.

\bibitem{anderson2018bottom}
P.~Anderson, X.~He, C.~Buehler, D.~Teney, M.~Johnson, S.~Gould, and L.~Zhang,
  ``Bottom-up and top-down attention for image captioning and visual question
  answering,'' in \emph{Proceedings of the IEEE Conference on Computer Vision
  and Pattern Recognition}, 2018.

\bibitem{cornia2019show}
M.~Cornia, L.~Baraldi, and R.~Cucchiara, ``{Show, Control and Tell: A Framework
  for Generating Controllable and Grounded Captions},'' in \emph{Proceedings of
  the IEEE Conference on Computer Vision and Pattern Recognition}, 2019.

\bibitem{vaswani2017attention}
A.~Vaswani, N.~Shazeer, N.~Parmar, J.~Uszkoreit, L.~Jones, A.~N. Gomez,
  {\L}.~Kaiser, and I.~Polosukhin, ``Attention is all you need,'' in
  \emph{Advances in Neural Information Processing Systems}, 2017.

\bibitem{herdade2019image}
S.~Herdade, A.~Kappeler, K.~Boakye, and J.~Soares, ``{Image Captioning:
  Transforming Objects into Words},'' in \emph{Advances in Neural Information
  Processing Systems}, 2019.

\bibitem{yao2018exploring}
T.~Yao, Y.~Pan, Y.~Li, and T.~Mei, ``{Exploring Visual Relationship for Image
  Captioning},'' in \emph{Proceedings of the European Conference on Computer
  Vision}, 2018.

\bibitem{jiang2018recurrent}
W.~Jiang, L.~Ma, Y.-G. Jiang, W.~Liu, and T.~Zhang, ``{Recurrent Fusion Network
  for Image Captioning},'' in \emph{Proceedings of the European Conference on
  Computer Vision}, 2018.

\bibitem{yao2010i2t}
B.~Z. Yao, X.~Yang, L.~Lin, M.~W. Lee, and S.-C. Zhu, ``I2t: Image parsing to
  text description,'' in \emph{Proceedings of the IEEE}, 2010.

\bibitem{socher2010connecting}
R.~Socher and L.~Fei-Fei, ``Connecting modalities: Semi-supervised segmentation
  and annotation of images using unaligned text corpora,'' in \emph{Proceedings
  of the IEEE Conference on Computer Vision and Pattern Recognition}, 2010.

\bibitem{liu2017improved}
S.~Liu, Z.~Zhu, N.~Ye, S.~Guadarrama, and K.~Murphy, ``Improved image
  captioning via policy gradient optimization of spider,'' in \emph{Proceedings
  of the International Conference on Computer Vision}, 2017.

\bibitem{johnson2016densecap}
J.~Johnson, A.~Karpathy, and L.~Fei-Fei, ``Densecap: Fully convolutional
  localization networks for dense captioning,'' in \emph{Proceedings of the
  IEEE Conference on Computer Vision and Pattern Recognition}, 2016.

\bibitem{vinyals2017show}
O.~Vinyals, A.~Toshev, S.~Bengio, and D.~Erhan, ``{Show and Tell: Lessons
  Learned from the 2015 MSCOCO Image Captioning Challenge},'' \emph{IEEE
  Transactions on Pattern Analysis and Machine Intelligence}, vol.~39, no.~4,
  pp. 652--663, 2017.

\bibitem{donahue2015long}
J.~Donahue, L.~Anne~Hendricks, S.~Guadarrama, M.~Rohrbach, S.~Venugopalan,
  K.~Saenko, and T.~Darrell, ``Long-term recurrent convolutional networks for
  visual recognition and description,'' in \emph{Proceedings of the IEEE
  Conference on Computer Vision and Pattern Recognition}, 2015.

\bibitem{rennie2017self}
S.~J. Rennie, E.~Marcheret, Y.~Mroueh, J.~Ross, and V.~Goel, ``Self-critical
  sequence training for image captioning,'' in \emph{Proceedings of the IEEE
  Conference on Computer Vision and Pattern Recognition}, 2017.

\bibitem{bahdanau2014neural}
D.~Bahdanau, K.~Cho, and Y.~Bengio, ``Neural machine translation by jointly
  learning to align and translate,'' in \emph{Proceedings of the International
  Conference on Learning Representations}, 2015.

\bibitem{xu2015show}
K.~Xu, J.~Ba, R.~Kiros, K.~Cho, A.~Courville, R.~Salakhutdinov, R.~S. Zemel,
  and Y.~Bengio, ``Show, attend and tell: Neural image caption generation with
  visual attention,'' in \emph{Proceedings of the International Conference on
  Machine Learning}, 2015.

\bibitem{you2016image}
Q.~You, H.~Jin, Z.~Wang, C.~Fang, and J.~Luo, ``Image captioning with semantic
  attention,'' in \emph{Proceedings of the IEEE Conference on Computer Vision
  and Pattern Recognition}, 2016.

\bibitem{cornia2018paying}
M.~Cornia, L.~Baraldi, G.~Serra, and R.~Cucchiara, ``{Paying more attention to
  saliency: Image captioning with saliency and context attention},'' \emph{ACM
  Transactions on Multimedia Computing, Communications, and Applications},
  vol.~14, no.~2, p.~48, 2018.

\bibitem{pedersoli2016areas}
M.~Pedersoli, T.~Lucas, C.~Schmid, and J.~Verbeek, ``Areas of attention for
  image captioning,'' in \emph{Proceedings of the International Conference on
  Computer Vision}, 2017.

\bibitem{ranzato2015sequence}
M.~Ranzato, S.~Chopra, M.~Auli, and W.~Zaremba, ``Sequence level training with
  recurrent neural networks,'' in \emph{Proceedings of the International
  Conference on Learning Representations}, 2015.

\bibitem{aneja2018convolutional}
J.~Aneja, A.~Deshpande, and A.~G. Schwing, ``Convolutional image captioning,''
  in \emph{Proceedings of the IEEE Conference on Computer Vision and Pattern
  Recognition}, 2018.

\bibitem{devlin2018bert}
J.~Devlin, M.-W. Chang, K.~Lee, and K.~Toutanova, ``{BERT: Pre-training of Deep
  Bidirectional Transformers for Language Understanding},'' in
  \emph{Proceedings of the Annual Conference of the North American Chapter of
  the Association for Computational Linguistics}, 2019.

\bibitem{he2016deep}
K.~He, X.~Zhang, S.~Ren, and J.~Sun, ``Deep residual learning for image
  recognition,'' in \emph{Proceedings of the IEEE Conference on Computer Vision
  and Pattern Recognition}, 2016.

\bibitem{ba2016layer}
J.~L. Ba, J.~R. Kiros, and G.~E. Hinton, ``{Layer Normalization},'' \emph{arXiv
  preprint arXiv:1607.06450}, 2016.

\bibitem{sukhbaatar2019augmenting}
S.~Sukhbaatar, E.~Grave, G.~Lample, H.~Jegou, and A.~Joulin, ``{Augmenting
  Self-attention with Persistent Memory},'' \emph{arXiv preprint
  arXiv:1907.01470}, 2019.

\bibitem{vedantam2015cider}
R.~Vedantam, C.~Lawrence~Zitnick, and D.~Parikh, ``{CIDEr: Consensus-based
  Image Description Evaluation},'' in \emph{Proceedings of the IEEE Conference
  on Computer Vision and Pattern Recognition}, 2015.

\bibitem{karpathy2015deep}
A.~Karpathy and L.~Fei-Fei, ``Deep visual-semantic alignments for generating
  image descriptions,'' in \emph{Proceedings of the IEEE Conference on Computer
  Vision and Pattern Recognition}, 2015.

\bibitem{hall2018probability}
D.~Hall, F.~Dayoub, J.~Skinner, H.~Zhang, D.~Miller, P.~Corke, G.~Carneiro,
  A.~Angelova, and N.~S{\"u}nderhauf, ``{Probabilistic Object Detection:
  Definition and Evaluation},'' \emph{arXiv preprint arXiv:1811.10800}, 2018.

\bibitem{papineni2002bleu}
K.~Papineni, S.~Roukos, T.~Ward, and W.-J. Zhu, ``{BLEU: a method for automatic
  evaluation of machine translation},'' in \emph{Proceedings of the Annual
  Meeting on Association for Computational Linguistics}, 2002.

\bibitem{lin2004rouge}
C.-Y. Lin, ``Rouge: A package for automatic evaluation of summaries,'' in
  \emph{Proceedings of the Annual Meeting on Association for Computational
  Linguistics Workshops}, 2004.

\bibitem{banerjee2005meteor}
S.~Banerjee and A.~Lavie, ``{METEOR: An automatic metric for MT evaluation with
  improved correlation with human judgments},'' in \emph{Proceedings of the
  Annual Meeting on Association for Computational Linguistics Workshops}, 2005.

\bibitem{kuhn1955hungarian}
H.~W. Kuhn, ``{The Hungarian method for the assignment problem},'' \emph{Naval
  Research Logistics Quarterly}, vol.~2, no. 1-2, pp. 83--97, 1955.

\bibitem{ren2015faster}
S.~Ren, K.~He, R.~Girshick, and J.~Sun, ``{Faster R-CNN: Towards real-time
  object detection with region proposal networks},'' in \emph{Advances in
  Neural Information Processing Systems}, 2015.

\bibitem{krishnavisualgenome}
R.~Krishna, Y.~Zhu, O.~Groth, J.~Johnson, K.~Hata, J.~Kravitz, S.~Chen,
  Y.~Kalantidis, L.-J. Li, D.~A. Shamma, M.~Bernstein, and L.~Fei-Fei,
  ``{Visual Genome: Connecting Language and Vision Using Crowdsourced Dense
  Image Annotations},'' \emph{International Journal of Computer Vision}, vol.
  123, no.~1, pp. 32--73, 2017.

\bibitem{pennington2014glove}
J.~Pennington, R.~Socher, and C.~Manning, ``{GloVe: Global vectors for word
  representation},'' in \emph{Proceedings of the Conference on Empirical
  Methods in Natural Language Processing}, 2014.

\bibitem{kingma2014adam}
D.~P. Kingma and J.~Ba, ``Adam: A method for stochastic optimization,'' in
  \emph{Proceedings of the International Conference on Learning
  Representations}, 2015.

\bibitem{yang2019auto}
X.~Yang, K.~Tang, H.~Zhang, and J.~Cai, ``{Auto-Encoding Scene Graphs for Image
  Captioning},'' in \emph{Proceedings of the IEEE Conference on Computer Vision
  and Pattern Recognition}, 2019.

\end{thebibliography}

\end{document}